\documentclass[10pt,twocolumn,letterpaper]{article}

\usepackage{cvpr}              
\definecolor{cvprblue}{rgb}{0.21,0.49,0.74}
\usepackage[pagebackref,breaklinks,colorlinks,allcolors=cvprblue]{hyperref}

\def\paperID{4654} 
\def\confName{CVPR}
\def\confYear{2026}

\newcommand{\ourmethod}{DTPTrack}
\usepackage{multirow}
\usepackage{makecell}
\newcommand*{\affaddr}[1]{#1} 
\newcommand*{\affmark}[1][*]{\textsuperscript{#1}}

\makeatletter
\def\thanks#1{\protected@xdef\@thanks{\@thanks\protect\footnotetext{#1}}}
\makeatother
\title{Drift-Resilient Temporal Priors for Visual Tracking}

\author{%
Yuqing Huang\affmark[1,2], Liting Lin\affmark[3], Weijun Zhuang\affmark[1,2], Zhenyu He\affmark[1,$\ast$], and Xin Li\affmark[2,4,$\ast$]\\
\affaddr{\affmark[1]Harbin Institute of Technology, Shenzhen}\quad
\affaddr{\affmark[2]Pengcheng Laboratory}\\
\affaddr{\affmark[3]Lero, the Research Ireland Centre for Software, University of Limerick}\quad
\affaddr{\affmark[4]Pazhou Lab (Huangpu)}
}
\begin{document}
\thanks{$\ast$ Corresponding author}
\maketitle
\begin{abstract}
Temporal information is crucial for visual tracking, but existing multi-frame trackers are vulnerable to model drift caused by naively aggregating noisy historical predictions. In this paper, we introduce \ourmethod{}, a lightweight and generalizable module designed to be seamlessly integrated into existing trackers to suppress drift. Our framework consists of two core components: (1) a \textbf{Temporal Reliability Calibrator (TRC)} mechanism that learns to assign a per-frame reliability score to historical states, filtering out noise while anchoring on the ground-truth template; and (2) a \textbf{Temporal Guidance Synthesizer (TGS)} module that synthesizes this calibrated history into a compact set of dynamic temporal priors to provide predictive guidance. To demonstrate its versatility, we integrate \ourmethod{} into three diverse tracking architectures—OSTrack, ODTrack, and LoRAT—and show consistent, significant performance gains across all baselines. Our best-performing model, built upon an extended LoRATv2 backbone, sets a new state-of-the-art on several benchmarks, achieving a 77.5\% Success rate on LaSOT and an 80.3\% AO on GOT-10k. The source code is available at https://github.com/NorahGreen/DTPTrack.

\end{abstract}
    
\section{Introduction}

\begin{figure*}[t]
    \centering
    \includegraphics[width=1.0\linewidth]{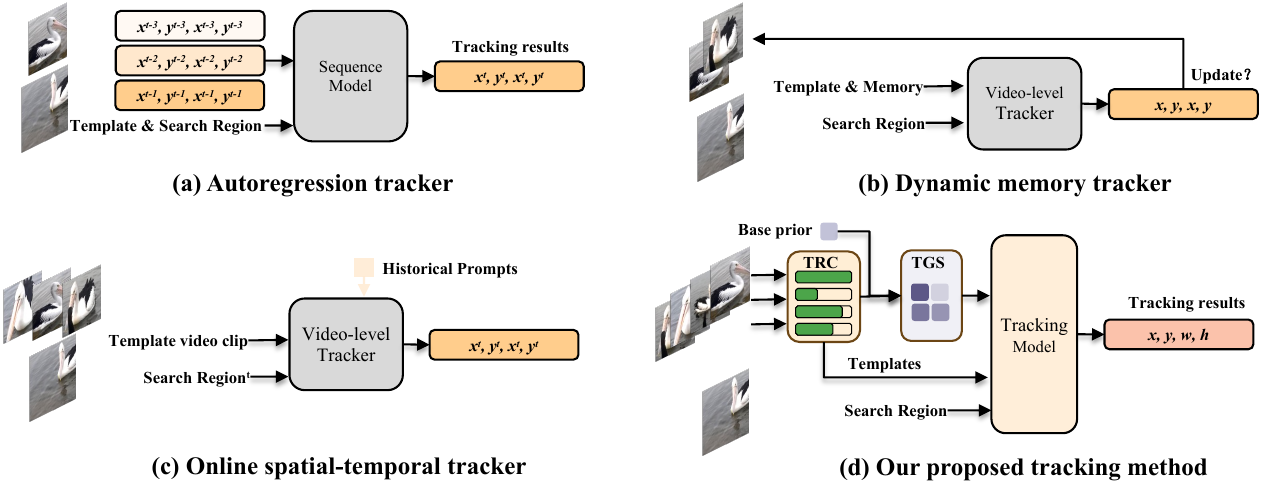}
    \caption{\textbf{Comparison of temporal modeling strategies in visual tracking.}
    (a) Autoregressive trackers propagate historical predictions through a sequence model, making them vulnerable to cumulative errors.
    (b) Dynamic memory trackers update an internal memory over time, but noisy predictions can contaminate the memory state.
    (c) Online spatial--temporal trackers process short video clips jointly but treat all historical frames as equally reliable.
    (d) \textbf{\ourmethod{}} uses Temporal Reliability Calibrator(TRC) to reliability-weight historical frames and Temporal Guidance Synthesizer (TGS) to produce temporal prior tokens that guide the tracking backbone and suppress drift.}

    \label{fig:motivation}
\end{figure*}

Visual Object Tracking (VOT) is a fundamental task in computer vision, with wide-ranging applications in autonomous driving, video surveillance, robotics, and human-computer interaction. The core challenge lies in accurately and robustly localizing a target specified in the initial frame throughout an entire video sequence. This requires models to be resilient to a host of real-world complexities, including severe occlusions, rapid motion, illumination changes, and significant variations in the target's appearance.

The advent of deep learning, and particularly the rise of the Vision Transformer (ViT) architecture~\cite{vit}, has catalyzed a paradigm shift in the field. Modern trackers such as OSTrack~\cite{ostrack}, MixFormer~\cite{mixformer}, and SeqTrack~\cite{seqtrack} have demonstrated remarkable performance by leveraging the power of self-attention to jointly learn target-specific features and model their relationship with the search region. A key driver of recent progress has been the incorporation of temporal information. Rather than relying solely on the initial template and the current search frame, state-of-the-art methods increasingly utilize a history of past frames to build a more comprehensive and dynamic model of the target. This is typically achieved through two main strategies: online model updating, where a dynamic template is progressively refreshed with recent predictions~\cite{stark, tatrack}, or direct multi-frame feature fusion, where features from several historical frames are jointly processed~\cite{seqtrack, odtrack}.

Despite these advances, a critical and persistent vulnerability plagues most temporal trackers: \textbf{model drift}. This phenomenon occurs when the tracker makes an inaccurate prediction—for instance, during a brief occlusion or due to a similar-looking distractor—and this erroneous information is subsequently "baked into" the target's temporal model. The corrupted model then leads to further inaccuracies in the next frame, initiating a cascade of compounding errors that often culminates in complete tracking failure. This ``Achilles' heel" prevents trackers from achieving the long-term robustness required for real-world applications. Both direct feature aggregation and naive template updating treat all historical information as equally valid, providing no mechanism to differentiate between a high-quality prediction and a noisy, corrupted one.

To address this fundamental challenge, we introduce \textbf{\ourmethod{}}, a novel, generalizable module designed to explicitly combat model drift by intelligently managing the flow of historical information. The key insight behind \ourmethod{} is that a robust temporal tracker must not only remember the past but also critically assess its reliability. Our module achieves this through a two-stage process. First, a \textbf{Temporal Reliability Calibrator (TRC)} mechanism learns to assign a per-frame reliability score to historical states, dynamically weighting the influence of each past prediction. It suppresses noise by down-weighting low-confidence frames. Second, a \textbf{Temporal Guidance Synthesizer (TGS)} module converts these calibrated historical summaries into a compact set of dynamic prior vectors. These vectors serve as a predictive cue, guiding the tracker's attention toward the target's likely next state without directly contaminating its raw visual feature processing.

A core design principle of \ourmethod{} is its generalizability. It is not a monolithic tracker but a lightweight, ``plug-and-play" module that can be seamlessly integrated into a wide range of existing architectures.
We validate this by integrating it into three distinct trackers: OSTrack~\cite{ostrack}, ODTrack~\cite{odtrack}, and LoRAT~\cite{lorat}, where it delivers consistent and significant performance improvements. When integrated into our primary backbone, an extended LoRATv2, our best model sets a new state-of-the-art on five challenging benchmarks, achieving a \textbf{77.5\%} success rate on LaSOT and an \textbf{80.3\%} AO on GOT-10k.

In summary, our main contributions are:
\begin{enumerate}
    \item We propose \textbf{\ourmethod{}}, a novel, plug-and-play module that explicitly gates historical information to suppress tracking drift. Its lightweight and self-contained design allows for easy integration into various existing trackers.
    
    \item We introduce a \textbf{Temporal Reliability Calibrator (TRC)} mechanism, which learns to assign a dynamic reliability score to each historical frame, adaptively filtering out noisy predictions.
    
    \item We design a \textbf{Temporal Guidance Synthesizer (TGS)} module that synthesizes the calibrated history into a compact set of dynamic prior tokens, providing effective temporal guidance to the host tracker without contaminating its visual features.
    
    \item We demonstrate the effectiveness and architectural independence of our approach through extensive experiments. \ourmethod{} consistently improves the performance of three diverse host trackers and achieves state-of-the-art results on five major tracking benchmarks.
\end{enumerate}
\section{Related Works}
\label{sec:related}

Our work builds upon recent advancements in Transformer-based tracking and directly addresses the long-standing challenge of temporal modeling and drift. In this section, we review the most relevant literature in these areas.

\subsection{Transformer-based Visual Tracking}
The introduction of the Transformer architecture~\cite{vit} has fundamentally reshaped the landscape of visual tracking~\cite{videotrack, droptrack, aiatrack, motiontrack, hiptrack, citetracker, ROMTrack, zhu2025two, diffusiontrack}. Early pioneering works like TransT~\cite{transt} and STARK~\cite{stark} adapted the original encoder-decoder structure for tracking. TransT utilized a dedicated feature fusion network to enhance template and search region interaction, while STARK proposed an end-to-end framework that directly predicted the target's bounding box, treating tracking as a direct regression problem. These models demonstrated the powerful context modeling capabilities of self-attention mechanisms.

Subsequent research~\cite{swintrack, artrack, tracker_grm, dreamtrack, chen2022high, rtracker} has focused on simplifying and strengthening these architectures. OSTrack~\cite{ostrack} and SimTrack~\cite{simtrack} eliminated the complex encoder-decoder design, showing that a plain ViT backbone could effectively perform feature extraction and interaction in a unified, one-stream manner. This simplified paradigm proved to be both efficient and highly effective. Other approaches, like MixFormer~\cite{mixformer} and MixViT~\cite{mixformer_extend}, have explored hybrid architectures that combine the local feature strengths of CNNs with the global context modeling of Transformers. Our work, \ourmethod{}, is not designed to replace these powerful backbones. Instead, it serves as a complementary module that can be integrated into them to specifically enhance their temporal reasoning and robustness against drift.

\subsection{Temporal Information Modeling in Tracking}
Leveraging information from past frames is a cornerstone of modern tracking. Existing approaches predominantly fall into two categories: online model updating and multi-frame feature fusion.

\noindent\textbf{Online Model Updating.} This classic strategy aims to adapt the tracker's model of the target over time. Early methods like MDNet~\cite{mdnet} fine-tuned a deep network online to handle appearance changes. More recent discriminative correlation filter (DCF) based trackers like DiMP~\cite{dimp} and PrDiMP~\cite{prdimp} learn a robust target model online to distinguish the target from the background. In the Transformer era, methods like TATrack~\cite{tatrack} have introduced dynamic update schemes where the initial template is progressively refreshed with information from recent, high-confidence predictions. The core challenge for all online updating methods is determining \emph{when} and with \emph{what} to update. An incorrect update, triggered by an occlusion or a distractor, can irreversibly corrupt the template, leading to immediate and catastrophic drift.

\noindent\textbf{Multi-Frame Feature Fusion.} An alternative to updating a single template is to directly process features from a sequence of historical frames. This approach allows the model to learn motion patterns and handle short-term occlusions by maintaining a richer temporal context. Trackers like STARK~\cite{stark}, ODTrack~\cite{odtrack}, and MCITrack~\cite{mcitrack}, and others~\cite{spmtrack, artrack, aqatrack}, concatenate feature maps from multiple past frames and feed them into a Transformer backbone for joint spatio-temporal modeling. While powerful, this strategy is also vulnerable to drift. It implicitly assumes that all historical frames are equally reliable. If one or more frames in the sequence contain inaccurate localizations, their features are fused without reservation, introducing noise that can mislead the tracker in the current frame.

\section{Method}
\label{sec:method}

In this section, we present the technical details of our \ourmethod{} module. The module is designed as a lightweight, plug-and-play component that can be inserted into a generic multi-frame visual tracker to enhance its temporal modeling capabilities and suppress drift. We first describe our primary tracking backbone, which is an extension of the powerful and efficient LoRATv2~\cite{loratv2} framework. We then provide a detailed breakdown of our proposed module, which consists of two core components: the Temporal Reliability Calibrator (TRC) and the Temporal Guidance Synthesizer (TGS). Finally, we explain how these components are seamlessly integrated into the backbone's information flow.

\begin{figure*}[t]
    \centering
    \includegraphics[width=0.78\linewidth]{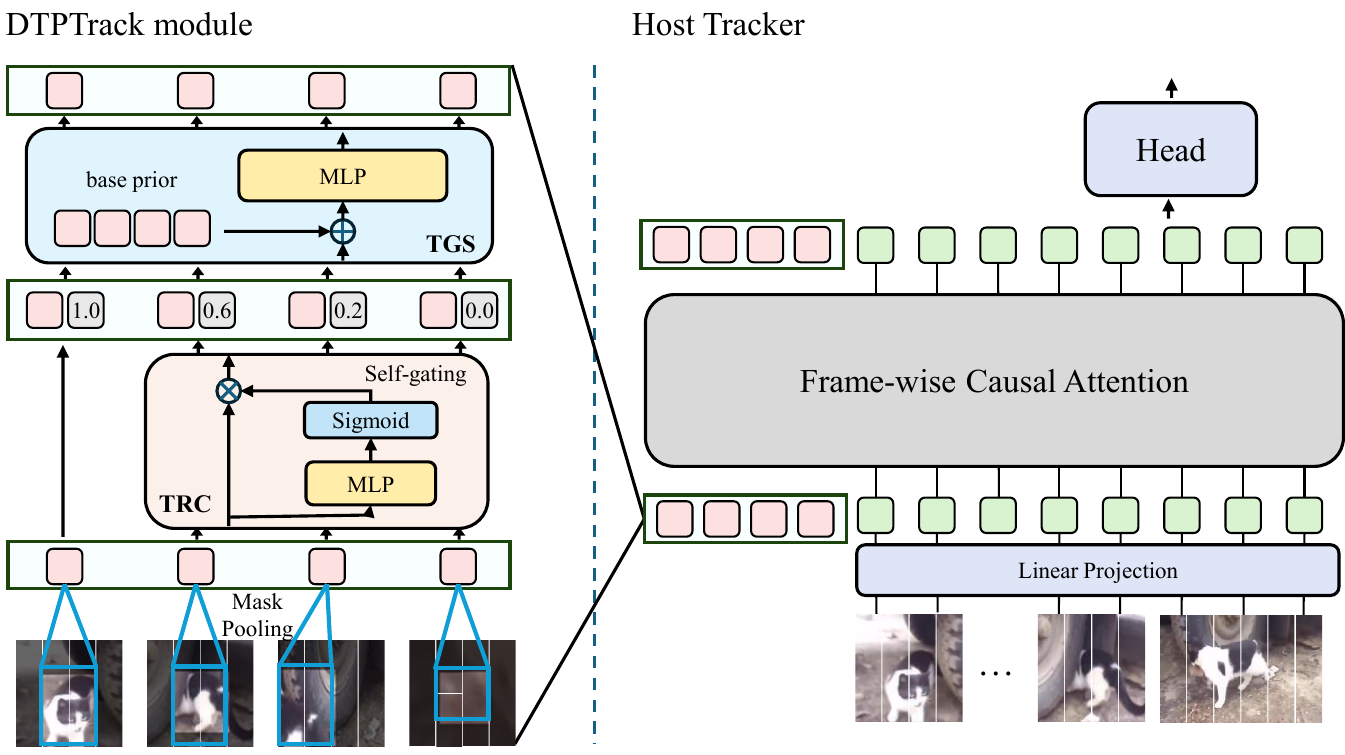}
    \caption{\textbf{Architectural Overview of the \ourmethod{} Module within our Extended LoRATv2 Backbone.} Our module operates in two stages before the main Transformer blocks. First, the \textbf{Temporal Reliability Calibrator (TRC)} block summarizes the feature embeddings of the template ($z_0$) and three historical reference frames ($z_1, z_2, z_3$) and computes a reliability weight for each. The confidence for the initial ground-truth frame ($z_0$) is fixed to 1.0. Second, the \textbf{Temporal Guidance Synthesizer (TGS)} block converts these calibrated summaries into a set of dynamic prior tokens. These tokens are then prepended to the original sequence of visual features to guide the host tracker's backbone through its Frame-Wise Causal Attention layers.}
    \label{fig:arch}
\end{figure*}

\subsection{Multi-Frame Tracking Backbone}

Our primary host tracker builds upon the LoRATv2 architecture, which is designed for efficient and effective temporal modeling. We adopt its core principles but extend it to process a five-frame sequence.

\noindent\textbf{LoRATv2 Core Principles.} We retain two key innovations from LoRATv2:
\begin{enumerate}
    \item \textbf{Frame-Wise Causal Attention (FWCA):} Standard self-attention has quadratic complexity with token count, making long sequences computationally expensive. FWCA addresses this by imposing a specific attention mask. Tokens within a given frame can attend fully to all other tokens \emph{within the same frame}, preserving rich spatial reasoning. However, they can only attend \emph{causally} to tokens in \emph{preceding} frames. This enforces a temporal autoregressive structure, significantly improving efficiency while effectively modeling sequential dependencies.
    \item \textbf{Stream-Specific LoRA Adapters (SSLA):} The causal structure introduces an asymmetry in how different input frames (streams) access historical information. To handle this, SSLA assigns a dedicated, lightweight Low-Rank Adaptation (LoRA)~\cite{lora} module to each input stream. This allows each stream to learn specialized adaptations (e.g., robust feature extraction for the template, motion-aware updates for reference frames) while sharing a single, frozen ViT backbone.
\end{enumerate}

\noindent\textbf{Backbone Extension.} Our specific implementation extends this backbone to process a sequence of five frames: an initial static template ($z_0$), three historical dynamic reference frames ($z_1, z_2, z_3$), and the current search region ($x_0$). The dynamic reference frames correspond to the search regions from the three preceding time steps. The full five-frame sequence is processed in each forward pass to ensure maximum representational flexibility.

\subsection{The \ourmethod{} Module for Drift Suppression}

While the LoRATv2 backbone provides an efficient structure for temporal modeling, it still implicitly treats all historical frames as equally reliable. To address this critical vulnerability, we introduce our \ourmethod{} module, which is composed of the TRC and TGS sub-components.

\subsubsection{Temporal Reliability Calibrator (TRC)}
The TRC module is tasked with a critical function: assessing the quality of the information provided by each historical frame ($z_0, z_1, z_2, z_3$).

First, to create an efficient and holistic representation of the target's state, we summarize the initial token embeddings from each frame. We operate on the features produced by the Vision Transformer's patch embedding layer, which performs a linear projection of the input image patches. For a given historical frame $i$, let $Z_i \in \mathbb{R}^{N_i \times D}$ be the sequence of these token embeddings. A corresponding binary mask, $M_i \in \{0, 1\}^{N_i}$, is generated, where a value of 1 indicates that a token's respective image patch intersects with the target object's bounding box. We then compute the summary vector $s_i \in \mathbb{R}^D$ via masked average pooling:
\begin{equation}
    s_i = \frac{\sum_{j=1}^{N_i} Z_{i,j} \cdot M_{i,j}}{\sum_{j=1}^{N_i} M_{i,j} + \epsilon}
    \label{eq:summary}
\end{equation}
This operation distills the essential appearance and location information of the target in frame $i$ into a single, dense vector.

Next, a confidence gate, implemented as a lightweight MLP with a sigmoid activation ($f_{\text{gate}}$), predicts a reliability score $c_i \in [0, 1]$ for each of the dynamic reference frames ($z_1, z_2, z_3$):
\begin{equation}
    [c_1, c_2, c_3] = \text{Sigmoid}(\text{MLP}( [s_1, s_2, s_3] ))
\end{equation}
Crucially, since the initial template $z_0$ is derived from ground truth, we \textbf{anchor} its confidence by fixing $c_0 = 1.0$. This ensures the temporal model always has a stable, uncorrupted reference, which we found to be vital for preventing long-term drift. The final calibrated summary vectors, $\hat{s}_i$, are obtained by element-wise multiplication: $\hat{s}_i = s_i \cdot c_i$.

\subsubsection{Temporal Guidance Synthesizer (TGS)}
The TGS module synthesizes the calibrated historical information into a format that can effectively guide the tracker's backbone. It converts the set of calibrated summary vectors $\{\hat{s}_0, \hat{s}_1, \hat{s}_2, \hat{s}_3\}$ into a small, fixed-size set of dynamic prior tokens.

The module begins with a set of learnable base prior tokens, $P_{\text{base}} \in \mathbb{R}^{K \times D}$, where $K$ is a small hyperparameter. A modulator MLP, $f_{\text{mod}}$, processes the sequence of calibrated summaries to produce a modulation signal, which is added to the base tokens to generate the final dynamic prior tokens, $P_{\text{dyn}}$:
\begin{equation}
    P_{\text{dyn}} = P_{\text{base}} + f_{\text{mod}}([\hat{s}_0, \hat{s}_1, \hat{s}_2, \hat{s}_3])
\end{equation}
These dynamic tokens are augmented with learnable positional and token-type embeddings to distinguish them from visual patch tokens.

\subsection{Integration with Host Tracker}


The integration of our module into a host tracker is designed for simplicity and broad compatibility. The general principle is to prepend the generated dynamic prior tokens to the tracker's standard input sequence:
\begin{equation}
    \text{Input} = \text{Concat}[P_{\text{dyn}}, Z_0, Z_1, \dots, X_0]
\end{equation}
Through the self-attention mechanism, these prior tokens provide high-level, drift-mitigated guidance to all visual tokens.

For trackers employing Frame-Wise Causal Attention (FWCA), such as LoRATv2~\cite{loratv2}, we make a specific adaptation: the prior tokens $P_{dyn}$ and the initial template tokens $Z_0$ are grouped together to form the \textbf{first computational chunk}. This establishes them as a stable, foundational context to which all subsequent frames can causally attend.

\section{Experiments}
\label{sec:experiments}

In this section, we conduct a comprehensive set of experiments to validate the effectiveness and generalizability of our proposed \ourmethod{} module. We first detail our experimental setup, including the host trackers, training data, and model configurations. We then present a thorough comparison against state-of-the-art methods on multiple challenging benchmarks. Finally, we will provide in-depth ablation studies to analyze the contribution of each component of our design.

\subsection{Experimental Setup}

\noindent\textbf{Model Variants.}
We provide two main variants of our tracker to demonstrate scalability:
\begin{itemize}
    \item \textbf{\ourmethod-B$_{224}$:} Utilizes a ViT-Base (DINOv2-B) backbone. The template and three reference frames have a resolution of $112 \times 112$, while the current search region is processed at $224 \times 224$ to capture finer details.
    \item \textbf{\ourmethod-L$_{378}$:} Utilizes a ViT-Large (DINOv2-L) backbone. The template and three reference frames have a resolution of $196 \times 196$, while the current search region is processed at a higher resolution of $378 \times 378$ to capture finer details.
\end{itemize}

\noindent\textbf{Implementation Details.}
All models are trained on a setup of four NVIDIA A100 GPUs, each with 40GB of memory. For our main results, we use an extended LoRATv2~\cite{loratv2} architecture as the host tracker, with the backbone ViT weights initialized from DINOv2~\cite{dinov2}. The backbone remains frozen during training; only our proposed \ourmethod{} module, the Stream-Specific LoRA (SSLA) adapters, and the prediction heads are trained. All other design and training settings (e.g., shared positional embeddings, anchor-free MLP heads, optimizer) follow the robust configuration of LoRATv2.

\noindent\textbf{Training.}
We follow the data pipeline established by recent state-of-the-art works like SPMTrack~\cite{spmtrack} and LoRAT~\cite{lorat}. The training set is a combination of LaSOT~\cite{lasot}, TrackingNet~\cite{trackingnet}, GOT-10k~\cite{got10k}, and COCO~\cite{coco}. During training, we sample 5-frame sequences. The first frame serves as the initial template ($z_0$), the subsequent three frames act as historical reference frames ($z_1, z_2, z_3$), and the final frame is the current search region ($x_0$). For evaluation on the GOT-10k test set, we strictly adhere to its protocol by training exclusively on the GOT-10k train split.

\noindent\textbf{Inference.}
During inference, we maintain a history of past predictions to populate the three dynamic reference frame slots. The selection strategy for these frames from the historical sequence is adapted from the methodology of SPMTrack~\cite{spmtrack}. We extend their approach to utilize three dynamic reference frames, compared to their two, to capture a richer temporal context. Additionally, following standard practice~\cite{ostrack, lorat, spmtrack}, we apply a Hanning window penalty to the final classification score map to suppress abrupt changes and improve localization stability.

\subsection{Comparison with State-of-the-Art Methods}

We evaluate our models on seven large-scale and challenging benchmarks: LaSOT~\cite{lasot}, VastTrack~\cite{vasttrack}, GOT-10k~\cite{got10k}, TrackingNet~\cite{trackingnet}, UAV123~\cite{uav123}, OTB2015~\cite{otb} and TNL2K~\cite{tnl2k}. We report standard metrics, including the Area Under the Curve (AUC) of success plots, normalized precision ($P_{Norm}$), precision ($P$), and Average Overlap (AO).
We compare our \ourmethod-B$_{224}$ and \ourmethod-L$_{378}$ models against a comprehensive suite of recent top-performing trackers. The detailed results are presented in Table~\ref{whole_comparison} and Table~\ref{table:uav,otb,tnl2k}.

\begin{table*}[!th]
    \centering
    \caption{State-of-the-art comparison on LaSOT, VastTrack, GOT-10k and TrackingNet. `*' denotes for trackers trained only with GOT-10k \emph{train} split. 
 The best three results are highlighted in \textbf{\textcolor{red}{red}}, \textbf{\textcolor{blue}{blue}} and \textbf{bold}, respectively.}\label{whole_comparison}
    \footnotesize
    \begin{tabular}{llccc cc ccc cc}
        \toprule
        \multirow{2.5}{*}{Method} & \multirow{2.5}{*}{Source}
        & \multicolumn{3}{c}{LaSOT}
        & \multicolumn{2}{c}{VastTrack}
        & \multicolumn{3}{c}{GOT-10k*}
        & \multicolumn{2}{c}{TrackingNet} \\
        \cmidrule(lr){3-5}\cmidrule(lr){6-7}\cmidrule(lr){8-10}\cmidrule(lr){11-12}
        & &
        AUC & P$_\textrm{Norm}$ & P
        & AUC & P
        & AO & SR$_{0.5}$ & SR$_{0.75}$
        & AUC & P$_\textrm{Norm}$ \\
        \midrule
        \ourmethod-B$_{224}$ & Ours & 74.3 & 83.6 & 80.7 & 40.7 & 40.2 & 76.5 & 87.6 & 74.9 & 85.6 & 90.1 \\
        \ourmethod-L$_{378}$ & Ours & \textbf{\textcolor{red}{77.5}} & \textbf{\textcolor{red}{86.5}} & \textbf{\textcolor{red}{84.7}} & \textbf{\textcolor{red}{47.2}} & \textbf{\textcolor{red}{49.7}} & \textbf{\textcolor{red}{80.3}} & \textbf{\textcolor{red}{89.9}} & \textbf{\textcolor{red}{81.9}} & \textbf{\textcolor{red}{86.9}} & \textbf{\textcolor{blue}{90.8}} \\
        \midrule
        LoRATv2-B$_{224}$~\cite{loratv2} & NIPS25 & 72.0 &81.3 &77.9 & 39.1 & 38.7 & 74.7 & 84.7 & 72.7 & 84.1 & - \\
        LoRATv2-L$_{378}$~\cite{loratv2} & NIPS25 & 76.1 & 85.1 & 83.1 & \textbf{44.2} & \textbf{46.7} & 78.2 & 86.8 & 79.1 & 85.7 & - \\
        SPMTrack-L~\cite{spmtrack} & CVPR25 & \textbf{\textcolor{blue}{76.8}} & \textbf{\textcolor{blue}{85.9}} & \textbf{\textcolor{blue}{84.0}} & - & - & \textbf{\textcolor{blue}{80.0}} & \textbf{\textcolor{blue}{89.4}} & \textbf{79.9} & \textbf{\textcolor{blue}{86.9}} & \textbf{\textcolor{red}{91.0}} \\
        LoRAT-B$_{378}$ \cite{lorat} & ECCV24 & 72.9 & 81.9 & 79.1 & 38.7 & 37.8 & 73.7 & 82.6 & 72.9 & 84.2 & 88.4 \\
        LoRAT-L$_{378}$ \cite{lorat} & ECCV24 & 75.1 & 84.1 & 82.0 & 43.9 & 45.8 & 77.5 & 86.2 & 78.1 & 85.6 & 89.7 \\
        LoRAT-g$_{378}$ \cite{lorat} & ECCV24 & \textbf{76.2} & \textbf{85.3} & \textbf{83.5} & \textbf{\textcolor{blue}{46.0}} & \textbf{\textcolor{blue}{48.8}} & 78.9 & 87.8 & \textbf{\textcolor{blue}{80.7}} & 86.0 & 90.2 \\
        AQATrack$_{384}$ \cite{aqatrack} & CVPR24 & 72.7 &82.9 & 80.2 & - & - & 76.0 & 85.2 & 74.9 & 84.8 & 89.3 \\
        ARTrackV2-B$_{384}$ \cite{artrackv2} & CVPR24 & 73.0 & 82.0 & 79.6 & - & - & 77.5 & 86.0 & 75.5 & 85.7 & 89.8 \\
        ARTrackV2-L$_{384}$ \cite{artrackv2} & CVPR24 & 73.6 & 82.8 & 81.1 & - & - & \textbf{79.5} & 87.8 & 79.6 & \textbf{86.1} & \textbf{90.4} \\
        HIPTrack \cite{hiptrack} & CVPR24 & 72.7 & 82.9 & 79.5 & - & - & 77.4 & \textbf{88.0}& 74.5 & 84.5 & 89.1 \\
        ODTrack-B \cite{odtrack} & AAAI24 & 73.2 & 83.2 & 80.6 & - & - & 77.0 & 87.9 & 75.1 & 85.1 & 90.1 \\
        ROMTrack$_{384}$ \cite{ROMTrack} & ICCV23 & 71.4 & 81.4 & 78.2 & 37.0 & 36.1 & 74.2 & 84.3 & 72.4 & 84.1 & 89.0 \\
        DropTrack$_{384}$  \cite{droptrack} & CVPR23 & 71.8 & 81.8 & 78.1 & 37.0 & 36.5 & 75.9 & 86.8 & 72.0 & 84.1 & 88.9 \\
        ARTrack-L$_{384}$  \cite{artrack} & CVPR23 & 73.1 & 82.2 & 80.3 & 35.6 & 32.4 & 78.5 & 87.4 & 77.8 & 85.6 & 89.6 \\
        SeqTrack-B$_{384}$   \cite{seqtrack} & CVPR23 & 71.5 & 81.1 & 77.8 & - & - & 74.5 & 84.3 & 71.4 & 83.9 & 88.8 \\
        GRM   \cite{tracker_grm} & CVPR23 & 69.9 & 79.3 & 75.8 & 36.3 & 34.8 & 73.4 & 82.9 & 70.4 & 84.0 & 88.7 \\
        TATrack-B  \cite{tatrack} & AAAI23 & 69.4 & 78.2 & 74.1 & - & - & 73.0 & 83.3 & 68.5 & 83.5 & 88.3 \\
        OSTrack$_{384}$  \cite{ostrack} & ECCV22 & 71.1 & 81.1 & 77.6 & 33.6 & 31.5 & 73.7 & 83.2 & 70.8 & 83.9 & 88.5 \\
        SimTrack  \cite{simtrack} & ECCV22 & 70.5 & 79.7 & - & - & - & 69.8 & 78.8 & 66.0 & 83.4 & 87.4 \\
        AiATrack  \cite{aiatrack} & ECCV22 & 69.0 & 79.4 & 73.8 & - & - & 69.6 & 80.0 & 63.2 & 82.7 & 87.8 \\
        \bottomrule
    \end{tabular} 
\end{table*}

\begin{table}[t]
    \centering
    \caption{Comparison with state-of-the-art methods on additional benchmarks in the AUC score.}\label{table:uav,otb,tnl2k}
    \footnotesize
    \begin{tabular}{l ccc}
    \toprule
        Method & UAV123 & OTB2015 & TNL2K \\
        \midrule
        {\ourmethod-L$_{378}$} & \textbf{72.3} & \textbf{74.7} & \textbf{63.7} \\
        {\ourmethod-B$_{224}$} & 71.9 & 73.2 & 60.3 \\
        SPMTrack-B & 71.7 & 72.7 & 62.0 \\
        HIPTrack \cite{hiptrack} & 70.5 & 71.0 & - \\
        ARTrackV2-B$_{256}$ \cite{artrackv2} & 69.9 & - & 59.2 \\
        ODTrack-B$_{384}$ \cite{odtrack} & - & 72.3 & 60.9 \\
        ARTrack$_{256}$  \cite{artrack} & 68.1 & - & 57.5 \\
        SeqTrack-B$_{384}$  \cite{seqtrack} & 68.6 & - & 56.4 \\
        DropTrack$_{384}$  \cite{droptrack} & - & 69.6 & 56.9 \\
    \bottomrule
    \end{tabular}
\end{table}

\noindent\textbf{LaSOT.} On this large-scale, long-term tracking benchmark, our \ourmethod-L$_{378}$ model achieves an AUC score of \textbf{\textcolor{red}{77.5\%}}, significantly outperforming all previous methods. This marks a substantial improvement over the already powerful SPMTrack-L (76.8\%) and LoRAT-g$_{378}$ (76.2\%), highlighting our method's exceptional robustness in long sequences where drift is a primary cause of failure. Similarly, our base model, \ourmethod-B$_{224}$, achieves an AUC of 74.3\%, surpassing its direct baseline LoRATv2-B$_{224}$ (72.0\%) by a large margin of 2.3 points.

\noindent\textbf{VastTrack.} This recently introduced benchmark features a vast number of object categories, testing a tracker's generalization capabilities. Our \ourmethod-L$_{378}$ model achieves a new state-of-the-art AUC of \textbf{\textcolor{red}{47.2\%}}. This result surpasses the previous best LoRAT-g$_{378}$ (46.0\%) and significantly improves upon the direct LoRATv2-L$_{378}$ baseline (44.2\%), demonstrating that our module enhances not just robustness but also generalization.

\noindent\textbf{GOT-10k.} On the GOT-10k benchmark, which enforces a strict protocol of training only on its designated training data, our method demonstrates strong generalization. \ourmethod-L$_{378}$ achieves a top-tier AO score of \textbf{\textcolor{red}{80.3\%}}, outperforming strong competitors like SPMTrack-L (80.0\%) and ARTrackV2-L (79.5\%). This indicates that our learned gating and prior generation mechanisms are effective even without exposure to massive, diverse pre-training datasets.

\noindent\textbf{TrackingNet.} On this large-scale short-term tracking benchmark, our \ourmethod-L$_{378}$ model is highly competitive, achieving an AUC of \textbf{\textcolor{red}{86.9\%}}, on par with the leading method SPMTrack-L. This shows that our approach provides its benefits without compromising performance on shorter, less complex sequences.

\noindent\textbf{UAV123.} This dataset focuses on challenging aerial drone scenarios. As shown in Table~\ref{table:uav,otb,tnl2k}, our \ourmethod-B$_{224}$ achieves a state-of-the-art AUC score of \textbf{\textcolor{red}{72.3\%}}, demonstrating its superior capability in handling small objects and complex motion in aerial views.

\noindent\textbf{OTB2015.} On the OTB2015 benchmark, our method again achieves a top-ranking AUC of \textbf{\textcolor{red}{74.7\%}}. This result validates its robustness against 11 common tracking challenges, including occlusion, deformation, and scale variation.

\noindent\textbf{TNL2K.} This benchmark is designed for language-guided tracking. Notably, our \ourmethod-B$_{224}$ achieves a competitive AUC of \textbf{\textcolor{red}{63.7\%}}, highlighting its strong performance.

The results clearly demonstrate the superiority of our proposed approach. Our \ourmethod{} module consistently elevates the performance of the strong LoRATv2 baseline, establishing new state-of-the-art results on several key benchmarks.


\subsection{Efficiency Comparison}

\begin{table}[t]
    \centering
    \caption{Efficiency comparison with state-of-the-art trackers. The number of frames processed per inference step is included to contextualize the computational cost. All measurements were conducted on a single NVIDIA A100 GPU. \# indicates number of frames incorporated. }
    \label{tab:efficiency_comparison}
    \footnotesize
    \begin{tabular}{l c ccc}
        \toprule
        Method & \#  & MACs (G) & Params (M) & FPS \\
        \midrule
        OSTrack-256~\cite{ostrack} & 2 & 21.5 & 92.1 & 106.2 \\
        OSTrack-384~\cite{ostrack} & 2 & 48.4 & 92.1 & 62.0 \\
        \midrule
        ODTrack-B~\cite{odtrack} & 5 & 73.1 & 92.1 & 40.5 \\
        ODTrack-L~\cite{odtrack} & 5 & 223.3 & 311.3 & 17.2 \\
        \midrule
        LoRAT-B$_{378}$~\cite{lorat} & 2 & 96.6 & 88.5 & 189.0 \\
        LoRAT-L$_{378}$~\cite{lorat} & 2 & 325.0 & 308.0 & 97.5 \\
        LoRAT-g$_{378}$~\cite{lorat} & 2 & 1161.0 & 1145.0 & 52.1 \\
        \midrule
        SPMTrack-B~\cite{spmtrack} & 4 & 293.0 & 88.5 & 10.0 \\
        SPMTrack-L~\cite{spmtrack} & 4 & 975.0 & 308.0 & 5.0 \\
        \midrule
        \textbf{\ourmethod-B$_{224}$} & 5 & 53.8 & 128.0 & 16.5 \\
        \textbf{\ourmethod-L$_{378}$} & 5 & 581.0 & 407.0 & 6.1 \\
        \bottomrule
    \end{tabular}
\end{table}

We analyze our method's computational profile in Table~\ref{tab:efficiency_comparison}, reporting MACs, parameter counts, and inference speed (FPS) benchmarked on a single NVIDIA A100 GPU.

The results highlight the computational advantage of our Frame-Wise Causal Attention (FWCA) backbone, which enables deeper temporal analysis with superior efficiency. For instance, our top-performing \textbf{\ourmethod-L$_{378}$} processes five frames yet requires fewer MACs than SPMTrack-L (581G vs. 975G), which only processes four. This demonstrates that FWCA provides a more scalable approach to multi-frame processing compared to methods relying on full spatio-temporal attention.

This efficiency allows us to deliberately prioritize accuracy by using a deep five-frame temporal modeling, which is reflected in our state-of-the-art results. Our base model, \textbf{\ourmethod-B$_{224}$}, offers a well-balanced profile, making it a viable option for scenarios with more constrained computational budgets. Overall, our method's computational profile reflects a strategic focus on maximizing accuracy through extensive temporal modeling, enabled by an efficient underlying architecture.

\subsection{Generalizability Analysis}

A core claim of our work is that \ourmethod{} is a versatile, plug-and-play module. To validate this, we integrated it into three representative trackers with different architectural and temporal modeling assumptions: OSTrack~\cite{ostrack}, a simple and popular two-frame tracker; ODTrack~\cite{odtrack}, a multi-frame detector-based tracker; and LoRAT~\cite{lorat}, an efficient two-frame model. For each tracker, we inject our module with minimal changes, training only the module's parameters while keeping the host backbone frozen. For a fair comparison, the OSTrack baseline we report is its official implementation with the Early Candidate Elimination (ECE) module disabled. The results on the LaSOT and VastTrack benchmarks are shown in Table~\ref{tab:generalization}.

\begin{table*}[t]
    \centering
    \caption{Generalizability analysis of \ourmethod{}. Our module consistently improves performance across three diverse host trackers with minimal computational overhead.}
    \label{tab:generalization}
    \footnotesize
    \begin{tabular}{l ccc cc cc}
        \toprule
        \multirow{2.5}{*}{Model} & \multirow{2.5}{*}{MACs(G)} & \multirow{2.5}{*}{Params(M)} & \multirow{2.5}{*}{FPS} & \multicolumn{2}{c}{LaSOT} & \multicolumn{2}{c}{VastTrack} \\
        \cmidrule(lr){5-6} \cmidrule(lr){7-8}
        & & & & AUC & P$_\textrm{Norm}$ & AUC & P \\
        \midrule
        OSTrack-256 & 29.05 & 92.12 & 118.5 & 69.1 & 78.7 & 33.6 & 31.5 \\
        + \ourmethod{} & 29.57 & 92.66 & 104.2 & \textbf{70.1} \textcolor{red!80!black}{(+1.0)} & \textbf{79.5} \textcolor{red!80!black}{(+0.8)} & \textbf{35.4} \textcolor{red!80!black}{(+1.8)} & \textbf{33.0} \textcolor{red!80!black}{(+1.5)} \\
        \midrule
        ODTrack-B & 73.10 & 92.12 & 40.5 & 73.2 & 83.2 & - & - \\
        + \ourmethod{} & 73.41 & 93.40 & 32.0 & \textbf{73.7} \textcolor{red!80!black}{(+0.5)} & \textbf{84.0} \textcolor{red!80!black}{(+0.8)} & \textbf{38.2} & \textbf{37.5} \\
        \midrule
        LoRAT-B224 & 33.24 & 98.71 & 176.0 & 71.1 & 80.9 & 38.7 & 37.8 \\
        + \ourmethod{} & 34.05 & 102.10 & 150.0 & \textbf{71.9} \textcolor{red!80!black}{(+0.8)} & \textbf{81.4} \textcolor{red!80!black}{(+0.5)} & \textbf{39.1} \textcolor{red!80!black}{(+0.4)} & \textbf{38.0} \textcolor{red!80!black}{(+0.2)} \\
        \bottomrule
    \end{tabular}
\end{table*}

As shown in Table~\ref{tab:generalization}, \ourmethod{} delivers consistent and notable performance improvements across all three host architectures, confirming its wide applicability.

\noindent\textbf{Consistent Performance Gains.} Our module successfully enhances trackers regardless of their design. It boosts the LaSOT AUC of the two-frame OSTrack by a significant \textbf{+1.0} point and the multi-frame ODTrack by \textbf{+0.5} points. The gains are particularly impressive on the challenging VastTrack benchmark, where OSTrack's AUC improves by a substantial \textbf{+1.8} points. This demonstrates that our explicit confidence gating and history synthesis provide a benefit that is orthogonal to the host's native temporal modeling strategy (or lack thereof).

\noindent\textbf{Minimal Computational Overhead.} The performance gains are achieved with remarkable efficiency. Integrating our module adds negligible computational cost: MACs increase by less than 1G for all trackers, and the number of trainable parameters increases by only 1-3M. The impact on practical FPS is also minimal, with all trackers maintaining high frame rates. This low overhead makes \ourmethod{} an attractive and practical addition to almost any tracking pipeline.

These results strongly validate our claim that \ourmethod{} is a truly generalizable and efficient module. It effectively enhances temporal robustness and overall performance, making it a valuable component for a wide range of visual trackers.

\subsection{Ablation Studies}

To validate our design choices and isolate the contributions of our key components, we conduct a series of ablation studies. We use our \ourmethod-B$_{224}$ model as the basis for these experiments, with all results reported on the LaSOT and VastTrack benchmarks. The comprehensive results are summarized in Table~\ref{tab:ablation_studies}.

\begin{table}[t]
    \centering
    \caption{Ablation studies on the core components of \ourmethod{}. All experiments are based on the ViT-Base backbone. The final row compares our full model against its direct multi-frame baseline.}
    \label{tab:ablation_studies}
    \footnotesize
    \begin{tabular}{l l cccc}
        \toprule
        \multirow{2.5}{*}{Variant} & \multirow{2.5}{*}{Modification} & \multicolumn{2}{c}{LaSOT} & \multicolumn{2}{c}{VastTrack} \\
        \cmidrule(lr){3-4} \cmidrule(lr){5-6}
        & & AUC & P$_\textrm{Norm}$ & AUC & P \\
        \midrule
        \multicolumn{6}{l}{\textit{Analysis of Temporal Reliability Calibrator (TRC)}} \\
        (a) & Fixed Threshold & 72.0 & 77.7 & 38.2 & 37.0 \\
        (b) & Fully Gated ($z_0$) & 73.2 & 79.3 & 40.1 & 39.4 \\
        \midrule
        \multicolumn{6}{l}{\textit{Analysis of Temporal Guidance Synthesizer (TGS)}} \\
        (c) & w/o Base Token & 72.7 & 78.4 & 39.0 & 38.2 \\
        (d) & Concat Fusion & 73.4 & 79.7 & 40.3 & 39.8 \\
        \midrule
        \multicolumn{6}{l}{\textit{Overall Module Effectiveness}} \\
        (e) & Baseline & 73.3 & 82.8 & 40.1 & 39.9 \\
        \textbf{(f)} & \textbf{Full Model} & \textbf{74.3} & \textbf{83.6} & \textbf{40.7} & \textbf{40.2} \\
        \bottomrule
    \end{tabular}
\end{table}

\noindent\textbf{Analysis of Temporal Reliability Calibrator (TRC).}
Our TRC component has two key design choices: a learned gate and an anchored initial frame. In variant (a), replacing our learned gate with a static, non-learned threshold results in a substantial performance drop of -2.3 AUC on LaSOT. This highlights the importance of learning to dynamically assess the reliability of historical frames. In variant (b), we allow the confidence gate to also score the initial frame ($z_0$) instead of fixing its confidence to 1.0. This modification also leads to a notable performance degradation, confirming our hypothesis that anchoring the model to a stable, fully-trusted ground-truth reference is crucial for preventing long-term drift.

\noindent\textbf{Analysis of Temporal Guidance Synthesizer (TGS).}
We analyze the two core aspects of our TGS module. In variant (c), we remove the learnable base priors and use the modulated signal directly. This harms performance, suggesting that the base tokens provide a stable, learnable foundation for the prior. In variant (d), we replace our decoupled prior token injection with a simpler feature concatenation of the summary vectors. The inferior performance indicates that our method of creating dedicated prior tokens is more effective, likely because it provides cleaner, high-level guidance without directly contaminating the raw visual features.

\noindent\textbf{Overall Module Effectiveness.}
We compare our full model (f) against its direct and strongest baseline (e): our extended LoRATv2 backbone. This baseline processes the same five-frame sequence but lacks our \ourmethod{} module. It is important to note that, unlike the original LoRATv2, neither of these models uses KV caching. This is because our adoption of the dynamic SPMTrack reference frame selection strategy makes pre-caching ineffective. As shown, our full model consistently outperforms this strong baseline, achieving a gain of \textbf{+1.0 AUC} on LaSOT. This result decisively demonstrates that the combined TRC and TGS components provide a significant and complementary benefit over the highly efficient multi-frame backbone alone.

\begin{table}[t]
\centering
\caption{Ablation study on the dynamic prior token generation method.}
\label{tab:ablation_prior}
\footnotesize
\begin{tabular}{l l cccc}
    \toprule
    \multirow{2.5}{*}{Variant} & \multirow{2.5}{*}{Modification} & \multicolumn{2}{c}{LaSOT} & \multicolumn{2}{c}{VastTrack} \\
    \cmidrule(lr){3-4} \cmidrule(lr){5-6}
    & & AUC & P$_\textrm{Norm}$ & AUC & P \\
    \midrule
    \multicolumn{6}{l}{\textit{Analysis of Dynamic Prior Token Generation}} \\
    (a) & Momentum-based & 73.8 & 80.3 & 40.1 & 39.5 \\
    (b) & Optical Flow-based & 73.2 & 79.3 & 39.4 & 38.7 \\
    \textbf{(c)} & \textbf{Ours} & \textbf{74.3} & \textbf{83.6} & \textbf{40.7} & \textbf{40.2} \\
    \bottomrule
\end{tabular}
\end{table}

\noindent\textbf{Analysis of Dynamic Prior Token Generation.}
We analyze our Temporal Guidance Synthesizer (TGS) by comparing it against two heuristic-based alternatives in Table~\ref{tab:ablation_prior}. Our proposed model (c) demonstrates a clear superiority over both the momentum-based (a) and optical flow-based (b) priors. The performance gap is particularly pronounced on the challenging VastTrack benchmark, which underscores the limitations of relying on fixed motion rules or noisy, low-level signals in complex scenarios. This validates our specific design choice: generating dynamic prior tokens by modulating a set of learnable base priors with a context-aware signal from an MLP is a fundamentally more robust and effective way to provide high-level guidance than using fixed heuristics.

\begin{table}[t]
\centering
\caption{Ablation study on the number of historical frames. We evaluate the model's performance by varying the temporal context length.}
\label{tab:ablation_frames}
\footnotesize
\begin{tabular}{l l cccc}
    \toprule
    \multirow{2.5}{*}{Variant} & \multirow{2.5}{*}{Modification} & \multicolumn{2}{c}{LaSOT} & \multicolumn{2}{c}{VastTrack} \\
    \cmidrule(lr){3-4} \cmidrule(lr){5-6}
    & & AUC & P$_\textrm{Norm}$ & AUC & P \\
    \midrule
    \multicolumn{6}{l}{\textit{Analysis of Temporal Context Length}} \\
    (a) & 2 Frames & 72.0 & 81.3 & 39.1 & 38.7 \\
    (b) & 3 Frames & 73.2 & 82.3 & 39.7 & 39.2 \\
    (c) & 4 Frames & 73.7 & 83.1 & 40.3 & 39.8 \\
    \textbf{(d)} & \textbf{5 Frames (Ours)} & \textbf{74.3} & \textbf{83.6} & \textbf{40.7} & \textbf{40.2} \\
    \bottomrule
\end{tabular}
\end{table}

\noindent\textbf{Analysis of Temporal Context Length.}
We also study the impact of the number of historical frames on performance. As shown in Table~\ref{tab:ablation_frames}, there is a clear and consistent improvement across all metrics on both LaSOT and VastTrack as we increase the temporal context from two (a) up to five frames (d). The model's performance peaks with five frames, our final configuration. This trend underscores the importance of a rich temporal context for robustly modeling target motion and handling appearance changes. We chose five frames as they achieve the best performance in our experiments, striking an effective balance between contextual richness and computational efficiency.

\section{Conclusion}
\label{sec:conclusion}

In this paper, we introduced \ourmethod{}, a novel, plug-and-play module designed to combat model drift in multi-frame visual trackers. By explicitly gating the reliability of historical information with our Temporal Reliability Calibrator (TRC) mechanism and synthesizing it into guidance cues with our Temporal Guidance Synthesizer (TGS) module, our method provides a principled approach to temporal modeling.
Extensive experiments show that our module not only lifts a strong baseline to state-of-the-art performance on LaSOT and GOT-10k, but also improves three diverse trackers, demonstrating broad applicability. Overall, it provides an effective and practical approach to enhancing temporal robustness in modern visual trackers.

\section{Acknowledgments}
This work was supported by the National Natural Science Foundation of China (Grant No.62476148) and the Guangdong Basic and Applied Basic Research Foundation (Grant No.2024A1515011292).
{
    \small
    \bibliographystyle{ieeenat_fullname}
    \bibliography{main}

@String(PAMI = {IEEE Trans. Pattern Anal. Mach. Intell.})

@String(CVPR= {IEEE Conf. Comput. Vis. Pattern Recog.})

@String(ICCV= {Int. Conf. Comput. Vis.})

@String(ECCV= {Eur. Conf. Comput. Vis.})

@String(NIPS= {Adv. Neural Inform. Process. Syst.})

@String(ICLR = {Int. Conf. Learn. Represent.})

@String(PAMI  = {IEEE TPAMI})

@String(CVPR  = {CVPR})

@String(ICCV  = {ICCV})

@String(ECCV  = {ECCV})

@String(NIPS  = {NeurIPS})

@String(ICLR  = {ICLR})

@String(AAAI = {AAAI})

@inproceedings{diffusiontrack,
  title={Diffusiontrack: Point set diffusion model for visual object tracking},
  author={Xie, Fei and Wang, Zhongdao and Ma, Chao},
  booktitle=CVPR,
  year={2024}
}

@inproceedings{zhu2025two,
  title={Two-stream beats one-stream: asymmetric siamese network for efficient visual tracking},
  author={Zhu, Jiawen and Tang, Huayi and Chen, Xin and Wang, Xinying and Wang, Dong and Lu, Huchuan},
  booktitle=AAAI,
  volume={39},
  number={10},
  year={2025}
}

@article{chen2022high,
  title={High-performance transformer tracking},
  author={Chen, Xin and Yan, Bin and Zhu, Jiawen and Lu, Huchuan and Ruan, Xiang and Wang, Dong},
  journal=PAMI,
  volume={45},
  number={7},
  pages={8507--8523},
  year={2022},
  publisher={IEEE}
}

@inproceedings{dreamtrack,
  title={DreamTrack: Dreaming the Future for Multimodal Visual Object Tracking},
  author={Guo, Mingzhe and Tan, Weiping and Ran, Wenyu and Jing, Liping and Zhang, Zhipeng},
  booktitle=CVPR,
  year={2025}
}

@inproceedings{dimp,
	title        = {Learning Discriminative Model Prediction for Tracking},
	author       = {Goutam Bhat and Martin Danelljan and Luc~Van Gool and Radu Timofte},
	year         = 2019,
	booktitle    = ICCV
}

@inproceedings{vit,
	title        = {An Image is Worth 16x16 Words: Transformers for Image Recognition at Scale},
	author       = {Dosovitskiy, Alexey and Beyer, Lucas and Kolesnikov, Alexander and Weissenborn, Dirk and Zhai, Xiaohua and Unterthiner, Thomas and Dehghani, Mostafa and Minderer, Matthias and Heigold, Georg and Gelly, Sylvain and Uszkoreit, Jakob and Houlsby, Neil},
	year         = 2021,
	booktitle    = ICLR
}

@article{got10k,
	title        = {GOT-10k: A Large High-Diversity Benchmark for Generic Object Tracking in the Wild},
	author       = {Huang, Lianghua and Zhao, Xin and Huang, Kaiqi},
    year={2021},
	journal      = {IEEE Transactions on Pattern Analysis and Machine Intelligence},
	publisher    = {IEEE},
	volume       = 43,
	number       = 5,
	pages        = {1562--1577}
}

@inproceedings{coco,
	title        = {Microsoft {COCO}: Common Objects in Context},
	author       = {Lin, Tsung-Yi and Maire, Michael and Belongie, Serge and Hays, James and Perona, Pietro and Ramanan, Deva and Doll{\'a}r, Piotr and Zitnick, C Lawrence},
	year         = 2014,
	booktitle    = ECCV
}

@inproceedings{uav123,
	title        = {A Benchmark and Simulator for UAV Tracking},
	author       = {Mueller, Matthias and Smith, Neil and Ghanem, Bernard},
	year         = 2016,
	booktitle    = ECCV
}

@inproceedings{trackingnet,
	title        = {{TrackingNet}: A Large-Scale Dataset and Benchmark for Object Tracking in the Wild},
	author       = {Muller, Matthias and Bibi, Adel and Giancola, Silvio and Alsubaihi, Salman and Ghanem, Bernard},
	year         = 2018,
	booktitle    = ECCV
}

@inproceedings{mdnet,
	title        = {Learning Multi-Domain Convolutional Neural Networks for Visual Tracking},
	author       = {Nam, Hyeonseob and Han, Bohyung},
	year         = 2016,
	booktitle    = CVPR
}

@inproceedings{giou,
	title        = {Generalized Intersection over Union: A Metric and A Loss for Bounding Box Regression},
	author       = {Rezatofighi, Hamid and Tsoi, Nathan and Gwak, JunYoung and Sadeghian, Amir and Reid, Ian and Savarese, Silvio},
	year         = 2019,
	booktitle    = CVPR
}

@article{otb,
	title        = {Object Tracking Benchmark},
	author       = {Wu, Yi and Lim, Jongwoo and Yang, Ming-Hsuan},
	year         = 2015,
	journal      = {IEEE Transactions on Pattern Analysis and Machine Intelligence},
	volume       = 37,
	number       = 9,
	pages        = {1834--1848}
}

@inproceedings{stark,
	title        = {Learning Spatio-Temporal Transformer for Visual Tracking},
	author       = {Yan, Bin and Peng, Houwen and Fu, Jianlong and Wang, Dong and Lu, Huchuan},
	year         = 2021,
	booktitle    = ICCV
}

@inproceedings{adamw,
	title        = {Decoupled Weight Decay Regularization},
	author       = {Loshchilov, Ilya and Hutter, Frank},
	year         = 2019,
	booktitle    = ICLR
}

@inproceedings{transt,
	title        = {Transformer Tracking},
	author       = {Chen, Xin and Yan, Bin and Zhu, Jiawen and Wang, Dong and Yang, Xiaoyun and Lu, Huchuan},
	year         = 2021,
	booktitle    = CVPR
}

@inproceedings{prdimp,
	title        = {Probabilistic Regression for Visual Tracking},
	author       = {Danelljan, Martin and Gool, Luc Van and Timofte, Radu},
	year         = 2020,
	booktitle    = CVPR
}

@inproceedings{tnl2k,
	title        = {Towards More Flexible and Accurate Object Tracking With Natural Language: Algorithms and Benchmark},
	author       = {Wang, Xiao and Shu, Xiujun and Zhang, Zhipeng and Jiang, Bo and Wang, Yaowei and Tian, Yonghong and Wu, Feng},
	year         = 2021,
	booktitle    = CVPR
}

@InProceedings{larsson2016fractalnet,
  title={{FractalNet}: Ultra-Deep Neural Networks without Residuals},
  author={Larsson, Gustav and Maire, Michael and Shakhnarovich, Gregory},
  booktitle =ICLR,
  year={2016}
}

@inproceedings{mixformer,
  title={{MixFormer}: End-to-End Tracking with Iterative Mixed Attention},
  author={Cui, Yutao and Jiang, Cheng and Wang, Limin and Wu, Gangshan},
  booktitle=CVPR,
  year={2022}
}

@inproceedings{aiatrack,
  title={{AiATrack}: Attention in Attention for Transformer Visual Tracking},
  author={Gao, Shenyuan and Zhou, Chunluan and Ma, Chao and Wang, Xinggang and Yuan, Junsong},
  booktitle=ECCV,
  year={2022}
}

@inproceedings{vot2022,
  title={The Tenth Visual Object Tracking VOT2022 Challenge Results},
  author={Kristan, Matej and Leonardis, Ale{\v{s}} and Matas, Ji{\v{r}}{\'\i} and Felsberg, Michael and Pflugfelder, Roman and K{\"a}m{\"a}r{\"a}inen, Joni-Kristian and Chang, Hyung Jin and Danelljan, Martin and Zajc, Luka {\v{C}}ehovin and Luke{\v{z}}i{\v{c}}, Alan and others},
  booktitle={ECCV Workshops},
  year={2022}
}

@inproceedings{ostrack,
  title={Joint Feature Learning and Relation Modeling for Tracking: A One-Stream Framework},
  author={Ye, Botao and Chang, Hong and Ma, Bingpeng and Shan, Shiguang and Chen, Xilin},
  booktitle=ECCV,
  year={2022}
}

@inproceedings{deit_3,
  title={{DeiT} {III}: Revenge of the {ViT}},
  author={Touvron, Hugo and Cord, Matthieu and J{\'e}gou, Herv{\'e}},
  booktitle=ECCV,
  year={2022},
  organization={Springer}
}

@inproceedings{dinov2,
  title={{DINOv2}: Learning Robust Visual Features without Supervision},
  author={Oquab, Maxime and Darcet, Timoth{\'e}e and Moutakanni, Th{\'e}o and Vo, Huy and Szafraniec, Marc and Khalidov, Vasil and Fernandez, Pierre and Haziza, Daniel and Massa, Francisco and El-Nouby, Alaaeldin and others},
  booktitle={TMLR},
  year={2024}
}

@inproceedings{swintrack,
 author = {Lin, Liting and Fan, Heng and Zhang, Zhipeng and Xu, Yong and Ling, Haibin},
 booktitle = NIPS,
 title = {{SwinTrack}: A Simple and Strong Baseline for Transformer Tracking},
 year = {2022}
}

@inproceedings{lasot,
  title={{LaSOT}: A High-Quality Benchmark for Large-Scale Single Object Tracking},
  author={Fan, Heng and Lin, Liting and Yang, Fan and Chu, Peng and Deng, Ge and Yu, Sijia and Bai, Hexin and Xu, Yong and Liao, Chunyuan and Ling, Haibin},
  booktitle=CVPR,
  year={2019}
}

@inproceedings{citetracker,
  title={{CiteTracker}: Correlating Image and Text for Visual Tracking},
  author={Li, Xin and Huang, Yuqing and He, Zhenyu and Wang, Yaowei and Lu, Huchuan and Yang, Ming-Hsuan},
  booktitle=ICCV,
  year={2023}
}

@inproceedings{artrack,
  title={Autoregressive Visual Tracking},
  author={Wei, Xing and Bai, Yifan and Zheng, Yongchao and Shi, Dahu and Gong, Yihong},
  booktitle=CVPR,
  year={2023}
}

@inproceedings{seqtrack,
    title = {{SeqTrack}: Sequence to Sequence Learning for Visual Object Tracking},
    author={Chen, Xin and Peng, Houwen and Wang, Dong and Lu, Huchuan and Hu, Han},
    booktitle = CVPR,
    year = {2023}
}

@inproceedings{droptrack,
  title={{DropMAE}: Masked Autoencoders with Spatial-Attention Dropout for Tracking Tasks},
  author={Wu, Qiangqiang and Yang, Tianyu and Liu, Ziquan and Wu, Baoyuan and Shan, Ying and Chan, Antoni B},
  booktitle=CVPR,
  year={2023}
}

@inproceedings{simtrack,
  title={Backbone is All Your Need: A Simplified Architecture for Visual Object Tracking},
  author={Chen, Boyu and Li, Peixia and Bai, Lei and Qiao, Lei and Shen, Qiuhong and Li, Bo and Gan, Weihao and Wu, Wei and Ouyang, Wanli},
  booktitle=ECCV,
  year={2022}
}

@inproceedings{lora,
title={Lo{RA}: Low-Rank Adaptation of Large Language Models},
author={Edward J Hu and Yelong Shen and Phillip Wallis and Zeyuan Allen-Zhu and Yuanzhi Li and Shean Wang and Lu Wang and Weizhu Chen},
booktitle=ICLR,
year={2022}
}

@article{mixformer_extend,
      title={{MixFormer}: End-to-End Tracking with Iterative Mixed Attention}, 
      author={Cui, Yutao and Jiang, Cheng and Wu, Gangshan and Wang, Limin},
      year={2024},
      pages={4129 - 4146},
      journal={IEEE Transactions on Pattern Analysis and Machine Intelligence}
}

@inproceedings{ROMTrack,
  title={Robust Object Modeling for Visual Tracking},
  author={Cai, Yidong and Liu, Jie and Tang, Jie and Wu, Gangshan},
  booktitle=ICCV,
  year={2023}
}

@inproceedings{tracker_grm,
  title={Generalized Relation Modeling for Transformer Tracking},
  author={Gao, Shenyuan and Zhou, Chunluan and Zhang, Jun},
  booktitle=CVPR,
  year={2023}
}

@article{memory_efficient_attn,
  title={Self-attention Does Not Need $O(n^{2})$ Memory},
  author={Rabe, Markus N and Staats, Charles},
  journal={arXiv preprint arXiv:2112.05682},
  year={2021}
}

@inproceedings{artrackv2,
  title={ARTrackV2: Prompting Autoregressive Tracker Where to Look and How to Describe},
  author={Bai, Yifan and Zhao, Zeyang and Gong, Yihong and Wei, Xing},
  booktitle=CVPR,
  year={2024}
}

@inproceedings{hiptrack,
  title={HIPTrack: Visual Tracking with Historical Prompts},
  author={Cai, Wenrui and Liu, Qingjie and Wang, Yunhong},
  booktitle=CVPR,
  year={2024}
}

@inproceedings{aqatrack,
  title={Autoregressive Queries for Adaptive Tracking with Spatio-Temporal Transformers},
  author={Xie, Jinxia and Zhong, Bineng and Mo, Zhiyi and Zhang, Shengping and Shi, Liangtao and Song, Shuxiang and Ji, Rongrong},
  booktitle=CVPR,
  year={2024}
}

@inproceedings{tatrack,
  title={Target-Aware Tracking with Long-term Context Attention},
  author={He, Kaijie and Zhang, Canlong and Xie, Sheng and Li, Zhixin and Wang, Zhiwen},
  booktitle=AAAI,
  year={2023}
}

@inproceedings{videotrack,
  title={VideoTrack: Learning to Track Objects via Video Transformer},
  author={Xie, Fei and Chu, Lei and Li, Jiahao and Lu, Yan and Ma, Chao},
  booktitle=CVPR,
  year={2023}
}

@inproceedings{rtracker,
  title={RTracker: Recoverable Tracking via PN Tree Structured Memory},
  author={Huang, Yuqing and Li, Xin and Zhou, Zikun and Wang, Yaowei and He, Zhenyu and Yang, Ming-Hsuan},
  booktitle=CVPR,
  year={2024}
}

@inproceedings{odtrack,
  title={ODTrack: Online Dense Temporal Token Learning for Visual Tracking},
  author={Zheng, Yaozong and Zhong, Bineng and Liang, Qihua and Mo, Zhiyi and Zhang, Shengping and Li, Xianxian},
  booktitle=AAAI,
  year={2024}
}

@inproceedings{lorat,
  title={Tracking Meets LoRA: Faster Training, Larger Model, Stronger Performance},
  author={Lin, Liting and Fan, Heng and Zhang, Zhipeng and Wang, Yaowei and Xu, Yong and Ling, Haibin},
  booktitle=ECCV,
  year={2024}
}

@inproceedings{vasttrack,
  title={VastTrack: Vast Category Visual Object Tracking},
  author={Peng, Liang and Gao, Junyuan and Liu, Xinran and Li, Weihong and Dong, Shaohua and Zhang, Zhipeng and Fan, Heng and Zhang, Libo},
  booktitle=NIPS,
  year={2024}
}

@inproceedings{dinov2_reg_token,
  title={Vision Transformers Need Registers},
  author={Darcet, Timoth{\'e}e and Oquab, Maxime and Mairal, Julien and Bojanowski, Piotr},
  booktitle=ICLR,
  year={2024}
}

@inproceedings{flashattention,
  title={Flash{A}ttention: Fast and Memory-Efficient Exact Attention with {IO}-Awareness},
  author={Dao, Tri and Fu, Daniel Y. and Ermon, Stefano and Rudra, Atri and R{\'e}, Christopher},
  booktitle=NIPS,
  year={2022}
}

@inproceedings{flashattention2,
  title={Flash{A}ttention-2: Faster Attention with Better Parallelism and Work Partitioning},
  author={Dao, Tri},
  booktitle=ICLR,
  year={2024}
}

@inproceedings{spmtrack,
  title={SPMTrack: Spatio-Temporal Parameter-Efficient Fine-Tuning with Mixture of Experts for Scalable Visual Tracking},
  author={Cai, Wenrui and Liu, Qingjie and Wang, Yunhong},
  booktitle=CVPR,
  year={2025}
}

@inproceedings{mcitrack,
  title={Exploring Enhanced Contextual Information for Video-Level Object Tracking},
  author={Kang, Ben and Chen, Xin and Lai, Simiao and Liu, Yang and Liu, Yi and Wang, Dong},
  booktitle=AAAI,
  year={2025}
}

@article{motiontrack,
  title = "MotionTrack: Learning Motion Predictor for Multiple Object Tracking",
  author = "Xiao, Changcheng and Cao, Qiong and Zhong, Yujie and Lan, Long and Zhang, Xiang and Luo, Zhigang and Tao, Dacheng",
  journal ="Neural Networks",
  volume = "179",
  pages = "106539",
  year = "2024",
  publisher ="Elsevier"
}

@inproceedings{loratv2,
  title={LoRATv2: Enabling Low-Cost Temporal Modeling in One-Stream Trackers},
  author={Lin, Liting and Fan, Heng and Zhang, Zhipeng and Huang, Yuqing and Wang, Yaowei and Xu, Yong and Ling, Haibin},
  booktitle="NeurIPS",
  year="2025"
}

@inproceedings{vots2024,
  title={The second visual object tracking segmentation VOTS2024 challenge results},
  author={Kristan, Matej and Matas, Ji{\v{r}}{\'\i} and Tokmakov, Pavel and Felsberg, Michael and Zajc, Luka {\v{C}}ehovin and Luke{\v{z}}i{\v{c}}, Alan and Tran, Khanh-Tung and Vu, Xuan-Son and Bj{\"o}rklund, Johanna and Chang, Hyung Jin and others},
  booktitle={ECCV Workshops},
  pages={357--383},
  year={2024},
  organization={Springer}
}
}

\clearpage
\setcounter{page}{1}
\maketitlesupplementary
\appendix

In this supplementary material, we provide more implementation details and visualizations for the proposed \ourmethod{}.

\section{Additional Implementation Details}
\label{sec:additional_details}
Our experimental setup largely aligns with the foundational configurations and hyperparameters established in LoRAT~\cite{lorat}. A summary of the essential hyperparameters, including both the training and inference phases, is provided in Tab.~\ref{tbl:hyperparameters}.

\begin{table}[h]
\caption{Hyper-parameters used in \ourmethod{}.}\label{tbl:hyperparameters}
\centering
\begin{tabular}{ l  c }
\toprule
Item & Value
\\
\midrule
template area factor & 2 \\
search region area factor & 4 (B-224) / 5 (L-378) \\
scale jitter & 0.25 \\
translation jitter & 3 \\
horizontal flip & 0.5 \\
color jitter & 0.4 \\
batch size & 128 \\
epochs & 170 / 100 (GOT-10k) \\
optimizer & AdamW \\
lr & 1e-4 \\
weight decay & 0.1 \\
drop path & 0.0 (B) / 0.1 (L) \\
clip max norm & 1.0 \\
lr\_min & 5e-6 \\
warmup epochs & 2 \\
warmup lr mult & 1e-3 \\
BCE loss coef & 1.0 \\
GIoU loss coef & 1.0 \\
Hann window penalty & 0.45 \\
LoRA rank $r$ & 64 \\
\bottomrule
\end{tabular}
\end{table}

\subsection{Model Configuration}
\label{sec:model_config}
We adopt standard ViT-B/14 and ViT-L/14 architectures~\cite{dinov2, vit} as backbones for our B-224 and L-378 models, respectively, initializing them with DINOv2 pre-trained weights~\cite{dinov2, dinov2_reg_token}. While the backbone parameters remain frozen, we introduce Low-Rank Adaptation (LoRA)~\cite{lora} with a rank of $r=64$ to all linear projection matrices within the attention and MLP blocks for efficient fine-tuning. Consistent with the methodology of LoRAT~\cite{lorat}, input images are tokenized using a patch size of 14 with shared positional embeddings, and the final prediction is handled by two separate 3-layer Multi-Layer Perceptrons (MLPs) for bounding box regression and classification.





\subsection{Training Details}
\label{sec:training}

\paragraph{Loss.}
The overall loss is a sum of a Binary Cross-Entropy (BCE) loss for classification and a GIoU loss~\cite{giou} for bounding box regression. We assign equal importance to both terms, setting their coefficients to 1.0.

\paragraph{Data Augmentation.}
We employ the common data augmentation pipeline~\cite{lorat} on the video clips, including random horizontal flipping (probability 0.5), color jittering (e.g., brightness, contrast, saturation), and the 3-Augment strategy~\cite{deit_3}. The template images are cropped with a 2× area factor around the initial bounding box or predicted boxes. Search regions are cropped with an area factor of 4 (for B-224) or 5 (for L-378). Scale jitter and translation jitter are also applied to both versions.

\paragraph{Optimization.}
Models are trained for 170 epochs (100 for GOT-10k) using AdamW~\cite{adamw} with a batch size of 128. We use an initial learning rate of $1 \times 10^{-4}$, a weight decay of 0.1 (excluding bias/norm), and gradient clipping at 1.0. The learning rate follows a cosine schedule decaying to $5 \times 10^{-6}$ after a 2-epoch linear warmup from $1 \times 10^{-7}$. For \ourmethod{}-L, a DropPath rate of 0.1 is applied~\cite{larsson2016fractalnet}. To maximize computation efficiency, we utilize Automatic Mixed Precision (AMP) and memory-efficient attention~\cite{memory_efficient_attn}.

\subsection{Inference Details}
\label{sec:inference}
During inference, a Hann window penalty is applied to the classification score map. The penalty coefficient is 0.45.
We employ optimized fused GPU kernels, such as those provided by FlashAttention~\cite{flashattention, flashattention2}, to reduce the latency incurred by attention-related memory operations and yield inference speeds (FPS) that more closely align with the theoretical FLOPs.

\section{More Experimental Results}
\label{sec:more_results}

In this section, we provide additional experimental results and detailed ablation studies to further demonstrate the effectiveness of our proposed \ourmethod{}.

\subsection{Attribute-wise Performance Analysis}
To understand how DTPTrack mitigates tracking drift, we evaluate the performance across different video attributes on the LaSOT benchmark. As shown in Table~\ref{tbl:attributes_lasot}, DTPTrack achieves consistent improvements across all categories, with the most significant gains observed in \textit{Scale Variation (SV)} (+2.3\%) and \textit{Aspect Ratio Change (ARC)} (+1.0\%), confirming that our temporal priors effectively handle geometric variations that typically lead to drift.

\begin{table}[h]
    \centering
    \caption{Attribute-wise AUC (\%) comparison on the LaSOT.}
    \label{tbl:attributes_lasot}
    \vspace{-2mm}
    \resizebox{\linewidth}{!}{
    \begin{tabular}{lcccccccc}
        \toprule
        Method & SV & ARC & ROT & POC & FOC & DEF & BC & LR \\
        \midrule
        Baseline & 71.8 & 71.8  & 73.0 & 70.8 & 65.6 & 74.7 & 66.8 & 66.8 \\
        \textbf{+ DTPTrack} & \textbf{74.1} & \textbf{72.8} & \textbf{73.5} & \textbf{71.5} & \textbf{66.7} & \textbf{75.0} & \textbf{68.2} & \textbf{68.4} \\
        \bottomrule
    \end{tabular}}
\end{table}

\subsection{Drift Rate Error (DRE) on DiDi}
We further quantify the drift reduction using the Drift Rate Error (DRE) metric on the DiDi dataset. As reported in Table~\ref{tbl:didi}, DTPTrack achieves a lower DRE compared to the baseline, indicating more stable long-term tracking performance.

\begin{table}[h]
    \centering
    \small
    \caption{Comparison of tracking stability and DRE on the DiDi dataset.}
    \label{tbl:didi}
    \vspace{-2mm}
    \begin{tabular}{lcccc}
        \toprule
        Method & Qual. & Acc. & Rob. & \textbf{DRE} $\downarrow$ \\
        \midrule
        Baseline & 0.668 & 0.742 & 0.809 & 0.100 \\
        \textbf{+ DTPTrack} & \textbf{0.680} & \textbf{0.760} & \textbf{0.840} & \textbf{0.080} \\
        \bottomrule
    \end{tabular}
\end{table}



\subsection{Additional Comparisons on VOT Benchmarks}
To further validate the generalization and robustness of DTPTrack, we conduct additional experiments on the VOT-STB2022~\cite{vot2022} and VOT2024~\cite{vots2024} benchmarks. We focus on bounding box (BBox) tracking to ensure a fair comparison regarding input resolution, MACs, and supervision level.

\begin{table}[h]
    \centering
    \caption{EAO performance comparison on VOT-STB2022 and VOT2024 benchmarks.}
    \label{tbl:vot_comparison}
    \vspace{-2mm}
    \resizebox{0.9\linewidth}{!}{
    \begin{tabular}{llcc}
        \toprule
        Benchmark & Method & Backbone & EAO $\uparrow$ \\
        \midrule
        \multirow{2}{*}{VOT-STB2022} & MixFormer-L & ViT-L & 0.582 \\
                                     & \textbf{DTPTrack-B$_{224}$} & ViT-B & \textbf{0.610} \\
        \midrule
        \multirow{2}{*}{VOTS2024}     & LoRAT-g & ViT-L & 0.536 \\
                                     & \textbf{DTPTrack-L$_{378}$} & ViT-L & \textbf{0.630} \\
        \bottomrule
    \end{tabular}}
\end{table}

As shown in Table~\ref{tbl:vot_comparison}, our DTPTrack-B$_{224}$ outperforms recent state-of-the-art methods like MixFormerL on the VOT-STB2022 benchmark with a smaller backbone. Furthermore, on the challenging VOTS2024 benchmark, our approach achieves a competitive EAO of 0.630, significantly surpassing previous BBox-based trackers like LoRAT-g. 


\begin{figure}[ht]
    \centering
    \includegraphics[width=0.85\linewidth]{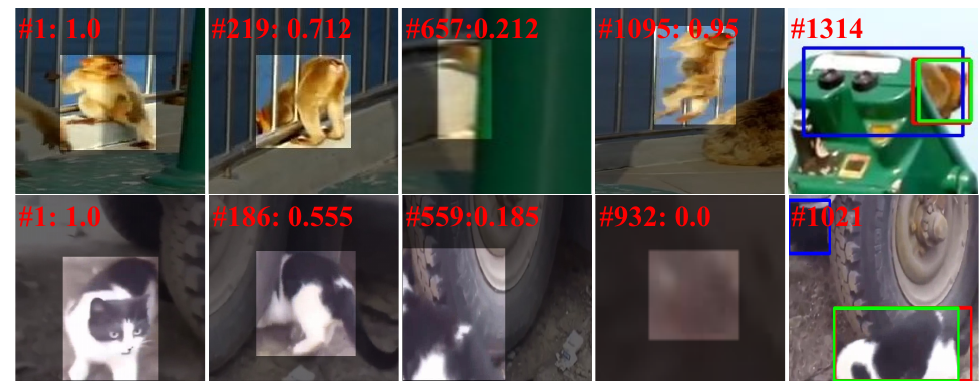}
    \caption{Visualization of the gating mechanism. The model assigns lower scores (indicated by darker colors or lower magnitude) to frames with artifacts or occlusions.}
    \label{fig:revised_viz}
\end{figure}

\begin{figure*}[ht]
    \centering
    \includegraphics[width=1.0\linewidth]{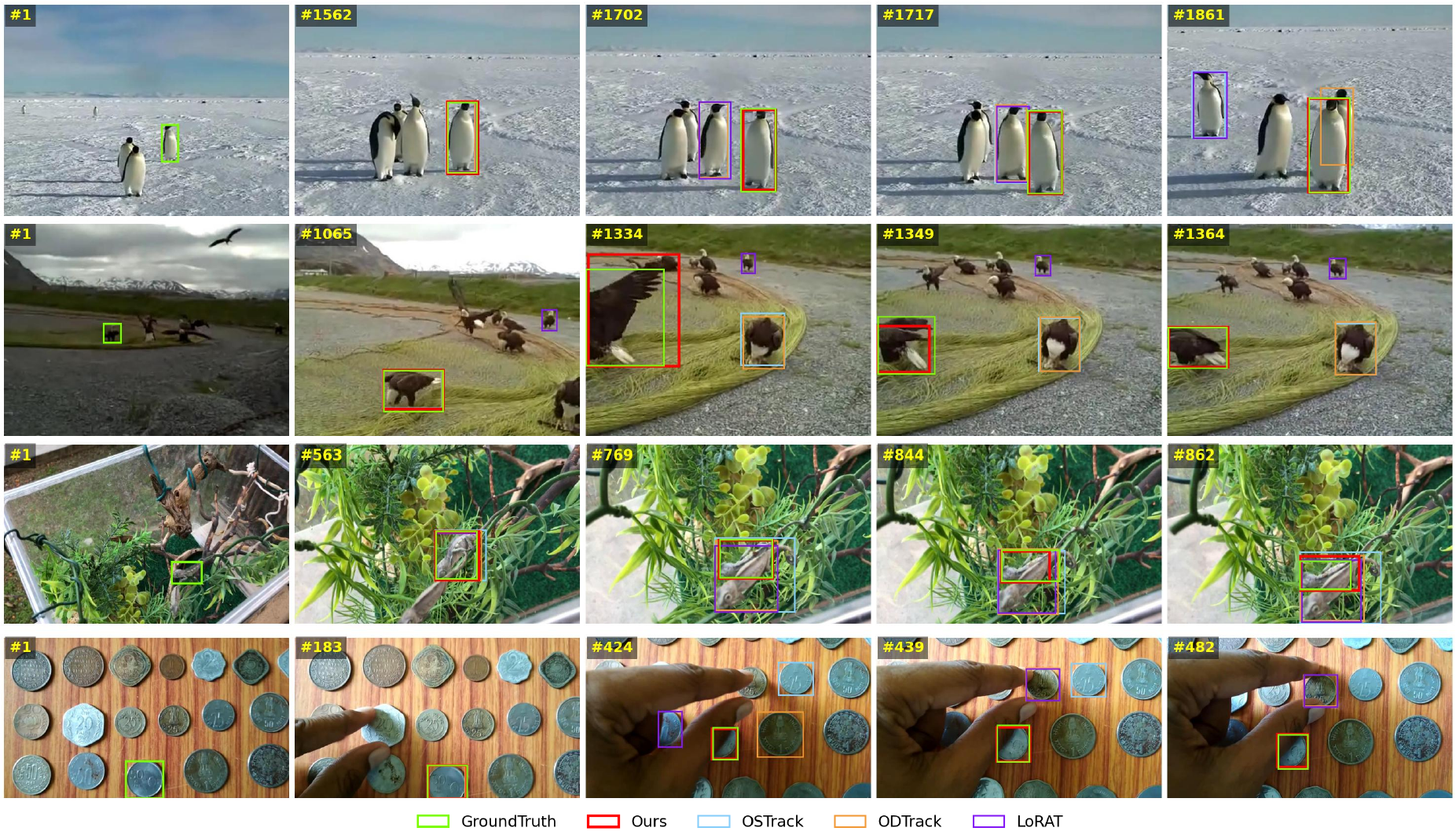}
    \caption{\textbf{Qualitative comparison with state-of-the-art trackers on challenging scenarios.}}
    \label{fig:sports_analysis}
\end{figure*}
\section{Visualization}
\subsection{Visualization of Gating Mechanism} 
Figure~\ref{fig:revised_viz} visualizes the learned gating values. It demonstrates that the model effectively assigns lower weights to frames with occlusion or low-quality visual features, successfully preventing noise from corrupting the temporal prior.

\subsection{Visualization of Challenging Sequences} 
In Fig.~\ref{fig:sports_analysis}, we provide qualitative comparisons between \ourmethod{} and three strong baselines (OSTrack, ODTrack, and LoRAT) on four challenging sequences. Green boxes denote ground-truth, while red, cyan, orange, and purple boxes represent the predictions of \ourmethod{}, OSTrack, ODTrack, and LoRAT, respectively. Across all examples, existing methods frequently drift to appearance-similar distractors, fall behind fast-moving targets, or lock onto background structures, resulting in fragmented or unstable trajectories.

In contrast, \ourmethod{} consistently stays tightly aligned with the target, even for small target sizes, heavy occlusion, strong background clutter, and dense distractor configurations. For instance, when multiple similar penguins appear in the scene, competing trackers gradually switch identities, whereas our tracker persists in tracking the correct instance. A similar pattern is observed in the bird (second row) and chameleon (third row) sequences, where high-textured background regions easily attract other methods, but \ourmethod{} maintains a stable prediction. 
In the coin-hand sequence with many near-duplicate objects, baselines often jump between coins or to the hand, while our predictions remain identity-consistent. These results visually confirm that the Temporal Reliability Calibrator and Temporal Guidance Synthesizer in \ourmethod{} effectively filter unreliable temporal cues and provide robust guidance to the backbone, leading to markedly improved long-term tracking stability.

\end{document}